\documentclass[conference]{IEEEtran}
\IEEEoverridecommandlockouts
\usepackage{cite}
\usepackage{amsmath,amssymb,amsfonts}
\usepackage{graphicx}
\usepackage{textcomp}
\usepackage{xcolor}
\usepackage{times}
\usepackage{xcolor}
\usepackage{todo}
\usepackage[small]{caption}
\usepackage{url}
\usepackage{hyperref}
\usepackage{algorithm}
\usepackage{algpseudocode}
\usepackage{booktabs}
\usepackage{makecell}
\usepackage{changes}

\begin{document}

\title{Improving Hearthstone AI by Combining {MCTS}\\ and Supervised Learning Algorithms\thanks{This research was co-funded by the Smart Growth Operational Programme 2014-2020, financed by the European Regional Development Fund under a GameINN projects POIR.01.02.00-00-0150/16 and POIR.01.02.00-00-0184/17, operated by The National Centre for Research and Development (NCBiR).}}

\author{\IEEEauthorblockN{Maciej \'Swiechowski\IEEEauthorrefmark{1}}
\IEEEauthorblockA{\IEEEauthorrefmark{1}Silver Bullet Labs\\
    Liwiecka 25, Warsaw, Poland\\
    {m.swiechowski@mini.pw.edu.pl}}
\and
\IEEEauthorblockN{Tomasz Tajmajer\IEEEauthorrefmark{1}\IEEEauthorrefmark{2}}
\IEEEauthorblockA{
	\IEEEauthorrefmark{2}Institute of Informatics, University of Warsaw\\
    Banacha 2, Warsaw, Poland\\
    {t.tajmajer@mimuw.edu.pl}
    }
\and
\IEEEauthorblockN{Andrzej Janusz\IEEEauthorrefmark{2}\IEEEauthorrefmark{3}}
\IEEEauthorblockA{
\IEEEauthorrefmark{3}eSensei \\
Mazowiecka 11/49, Warsaw, Poland \\
{janusza@mimuw.edu.pl}}
}

\maketitle

\added{Warning: this is not the final camera-ready version.}
%
%
\begin{abstract}
We investigate the impact of supervised prediction models on the strength and efficiency of artificial agents that use the Monte-Carlo Tree Search (MCTS) algorithm to play a popular video game {\it Hearthstone: Heroes of Warcraft}. We overview our custom implementation of the MCTS that is well-suited for games with partially hidden information and random effects. We also describe experiments which we designed to quantify the performance of our Hearthstone agent's decision making. We show that even simple neural networks can be trained and successfully used for the evaluation of game states. Moreover, we demonstrate that by providing a guidance to the game state search heuristic, it is possible to substantially improve the win rate, and at the same time reduce the required computations. 
\end{abstract}


\section{Introduction}\label{sec:intro}

{\it Hearthstone: Heroes of Warcraft} is a free-to-play online video game developed and published by Blizzard Entertainment. 
Its simple rules and appealing design made this game successful among casual players. According to Blizzard's data, in 2017 the player-base of the game was about 70 million and it grows with each of the released expansions. The game is also popular within the eSport community, with cash-prize tournaments and many international events every year.

Hearthstone is an example of a turn-based collectible card game. During the game, two players choose their hero with a unique power and compose a deck of thirty cards. They spend mana points to cast spells, weapons and summon minions to attack the opponent, with the goal to reduce the opponent's health to zero or below. 
Due to a large number of distinct cards which implement various game mechanics, and special in-game effects which often have randomized outcomes, Hearthstone is an example of a game where actions may have non-deterministic results. Moreover, during a game each player is unaware of cards that the opponent holds in hand, nor the ordering of yet-to-be-drawn cards in his deck.
Finally, since a player may perform several actions in each turn of the game and ordering of those actions is pivotal to player's success, Hearthstone features great combinatoric complexity. 
All the above properties make Hearthstone a demanding challenge for AI-controlled bots that are designed to play this game. One objective of this article is to explain how our implementation of the Monte Carlo Tree Search (MCTS) algorithm deals with those problems. We also aim to discuss the means by which MCTS can be facilitated by machine learning algorithms and provide experimental evaluation of its performance.

The paper is organized as follows. In the next section, we continue with providing context of the research and show related initiatives. In Section~\ref{sec:mcts}, the MCTS algorithm is discussed with the focus on problems encountered in Hearthstone such as randomness, hidden information and combinatorial complexity. We also shed some light on the game simulator used for this research. The subsequent section is devoted to methods of combining MCTS with machine-learning-based heuristics. Finally, the last two sections contain a description of empirical experiments which we conducted to evaluate our Hearthstone agents and conclusions, respectively.



\section{Related work}\label{sec:relatedwork}

In recent years, Hearthstone has become a testbed for AI research. A community of passionate players and developers have started the HearthSim project (\url{https://hearthsim.info/}) and created several applications that allow simulating the game for the purpose of AI and machine learning experiments. A few spin-offs of that project, e.g. \href{http://www.hearthpwn.com/}{HearthPWN} and \href{http://metastats.net/}{MetaStats}, provide tools for the players, which facilitate gathering data from their games. These portals obtain and aggregate users' data, such as game results, deck compositions, card usage statistics and provide this information to the community.

Several groups of researchers from the field of machine learning and AI have already chosen Hearthstone for their studies. In~\cite{garcia2016evolutionary}, authors used evolutionary algorithms to tackle the problem of building good decks. They used the results of simulated games performed by simple AI bots as fitness function values. Even though this study was described by the authors as preliminary, the developed method was able to construct reasonable decks from a basic set of cards. However, one drawback of this method is the fact that it strongly depends on the performance of the AI bots used for the evaluation of the decks.

A few research groups were also considering a problem of constructing an artificial agent able to play Hearthstone. In particular, \cite{DBLP:conf/cig/SantosSM17} used Monte-Carlo Tree Search (MCTS) algorithm to choose an optimal action policy in the game. Furthermore, \cite{DBLP:conf/cig/ZhangB17} used deep neural networks to improve performance of a MCTS-based Hearthstone bot, called Silverfish.
The combination of MCTS with prediction models make those approaches similar to early versions of DeepMind's AlphaGo program \cite{alphaGo}. It is worth noticing, however, that unlike Go, in Hearthstone players do not have full information about the game state and many actions have non-deterministic outcomes. These two properties make this game much more challenging for the game state tree search algorithms, such as MCTS \cite{6203567}. 

There were also attempts at constructing models for predicting cards that are likely to be played by an opponent during a game. For instance, in \cite{DBLP:conf/cig/Bursztein16} the author used data from 45,000 Hearthstone games to extract sequences of played cards and represent each record as a bag of card bi-grams. By investigating co-occurrence probabilities, the method described in that study was able to correctly predict opponent's card which will most likely appear during the following turns of the game, in over 50\% of cases. Such a high predictability can be explained by the fact that even though the number of possible Hearthstone decks is enormous, players tend to build their decks in accordance to certain archetypes and their composition is often inspired by the decks used by other influential players.

Hearthstone was also a topic of international data mining competitions. The first one, \textit{AAIA'17 Data Mining Challenge: Helping AI to Play Hearthstone}\footnote{Competition's web page: \url{https://knowledgepit.fedcsis.org/contest/view.php?id=120}}, was focused on developing a scoring model for predicting win chances of a player, based on detailed description of a single game state \cite{fedcsis_JanuszTS17}. Although the data in this competition was generated using very simple bots which were choosing their moves at random, the best models created by participants were able to achieve AUC scores above $0.80$. The winner used an ensemble of 1-dimensional convolutional neural networks to extract features from each combination of both players' cards on the board \cite{fedcsis/Grad17}. A year later, the second edition of this challenge was launched. The task in \textit{AAIA'18 Data Mining Challenge}
was to predict win-rates of Hearthstone decks, based on a history of match-ups between AI bots playing with similar decks.

Various other card games were also studied in the literature related to machine learning and AI. For instance, in \cite{Moravcik508} authors consider heads-up no-limit poker as an example of a game with hidden information. They describe a DeepStack algorithm which aims to handle the information asymmetry between players by combining recursive reasoning with learning from self-played games. As a different example one can give the game Magic: The Gathering, studied, e.g. in \cite{journals/tciaig/CowlingWP12}. Due to the notable similarity to Hearthstone, these games pose many similar challenges. In our work, however, we focus  only on Hearthstone.
The growing interest of the machine learning community in applications related to video games stems from the fact that solutions to many game-related problems could be easily transfered to real-life issues, such as planning~\cite{munos2014bandits}, real-time decision making~\cite{lee2008game,buro2003rts} and, ultimately, general~AI. 

\section{Playing Hearthstone with Monte-Carlo Tree Search}\label{sec:mcts}

\subsection{Game Simulator}

The access to a game simulator allows game-playing agents to perform dynamic reasoning about the game. The idea is to run separate simulations that do not affect the actual (main) state of the played game. This is a reason why a simulator is often called a ``forward model'' as it enables forward planning. Its performance, i.e., how many states it can visit per second, is crucial for all methods that are based on searching the space of the game such as MCTS, min-max or MTD(f). Therefore, we have written a simulator for Hearthstone with the aim of achieving the highest run-time performance. 
The main features of our simulator are: (1) written entirely in C++ for high-performance, (2) it performs 10K full games per second, in average, and 30K when limiting to basic cards only, (4) makes big use of inheritance and polymorphism (e.g., Secret : Spell : Card), (5) effects such as hero powers are modeled as (non-collectible) cards, (6) the total number of implemented cards = 483, (7) the implemented cards allow for making staple decks from the standard meta-game. 

The simulator calculates legal moves in each state of the game, updates the state after a move is chosen, tests whether the game reached a terminal state and calculates scores in a finished game. States and actions are comparable and hashable. We have divided complex game actions into atomic simple actions, e.g., when the ``SI-7 Agent'' card is played, up to three simple actions are generated: (1) Choose a card from your hand (SI-7 Agent), (2) Choose a target on the battle-field, where the minion is about to be placed, (3) Choose a target for the battle-cry: deal 2 damage, provided that the required \textit{combo} condition was met. Similarly, an attack move consists of two simple actions - choosing a character, which will attack and choosing a target to attack.

\subsection{Monte Carlo Tree Search}

Monte-Carlo Tree Search (MCTS)~\cite{mctsSurvey} has become the state-of-the-art algorithm for game tree search. It is the algorithm to go in domains such as Go~\cite{gellyGo}, Hex~\cite{hex}, Arimaa~\cite{arimaa},  General Game Playing (GGP)~\cite{ggp} or General Video Game Playing (GVGP)~\cite{gvgp}. This technique is a natural candidate for universal domains such as GGP or GVGP, because given only the way (interface) to simulate games, the same implementation of MCTS will work for any game. It has also been increasingly successful in board games such as Settlers of Catan~\cite{catan} or 7 Wonders~\cite{seven_wonders}.

In essence, the MCTS is a combination of three ideas: storing statistics in the game tree, random sampling by means of simulations to gather statistics and the Upper Confidence Bounds method to select nodes based on the statistics gathered so far. The \textit{Upper Confidence Bounds applied to Trees (UCT)} addresses the exploitation-exploration problem and it is a generalization of the Upper Confidence Bounds (UCB-1) method. The UCT formula is as follows:
\begin{equation}
\label{eq:uct}
a^{*} = \arg\max_{a \in A(s)}\left \{ Q(s,a)+C\sqrt{\frac{ln\left
[N(s)  \right ]}{N(s,a)}} \right \}
\end{equation}
where $A(s)$ is a set of actions available in state $s$, $Q(s,a)$ denotes the average result of playing action $a$ in state $s$ in the simulations performed so far, $N(s)$ - a number of times state $s$ has been visited in previous simulations and $N(s,a)$ - a number of times action $a$ has been sampled in this state in previous simulations. Constant $C$ controls the balance between exploration and exploitation. It has to be tuned, but provided that scores of games are confined to the $[0,1]$ interval, the sensible starting value is $\sqrt{2}$.  

The algorithm typically consists of four phases: selection, expansion, simulation and backpropagation. Algorithms (1) and (2) describe the usage of these phases.

\textbf{(1) Selection.} Traverse the nodes, that are already stored in the tree. At each level, the next node is chosen according to the selection policy - the UCT method, by default. 
\\
\indent \textbf{(2) Expansion.} A certain number of new nodes is added to the tree. In the classical MCTS variant, only one node is added by each iteration, which is a good trade-off between the algorithm's efficiency and memory usage.\\
\indent \textbf{(3) Simulation.} Starting from the last visited state in the tree, play (simulate) the game till the end. No nodes are added to the tree in this phase. Actions for each player are chosen randomly, however, there are extensions of the MCTS algorithm that introduce heuristics in the simulation. This phase is also called ``Monte-Carlo phase''.\\
\indent \textbf{(4) Back-propagation.} Starting from the last visited node in the tree, which is the one the simulation started from, all the way up to the root node, update the $Q(s,a)$ values based on the result of the simulation.


\subsubsection{Handling Imperfect Information}

The majority of successful applications of the MCTS algorithm have been done in the realm of perfect information games, i.e., games in which each player has complete information about the current state of the game. Games with hidden information have been proven to be difficult for any combinatorial method such as game-tree search. 
There have been many variants and extensions to the MCTS proposed to deal with imperfect information. However, they can be clustered into two types of approaches:
\begin{enumerate}
\item \textbf{Perfect Information Monte Carlo Tree Search (PIMC)} - this method determines (guesses) all information that is hidden and, from that point, treats the game as perfect information one. Variants of PIMC differ in the way how many distinct determinizations they perform and how the knowledge obtained from running the algorithm with different determinizations is combined. 
The two major problems related to PIMC~\cite{pimc} are strategy fusion and nonlocality~\cite{cowling2012information}.
\item \textbf{Information Set Monte Carlo Tree Search (ISMCTS)}~\cite{cowling2012information} - this variant uses the concept of information sets, which are abstract groups of states that are indistinguishable from a particular player's perspective. In ISMCTS, a node in the game tree is associated with an information set rather than a single state. Therefore, the decisions of a player are made based upon what the player actually observes. ISMCTS is much less susceptible to the problems of strategy fusion and nonlocality. However, ISMCTS is typically much harder to implement as it requires to simulate games under imperfect information or deal with partially observable moves. 
\end{enumerate}

We propose an algorithm, which is a combination of ISMCTS and PIMC.
From the first concept, we borrow the idea of information sets. However, they are not used to simulate games under hidden information. Instead, they serve as keys in the so-called transposition table. The transposition tables are a way to model the ``game-tree'' without duplicated nodes, which would occur if there is more than one way to reach the same state. The ``tree'' effectively then becomes a directed acyclic graph (DAG). Transposition tables are also often used to combine symmetric states in order to reuse calculations. In the transposition table we used, the values are nodes and there is a unique key-value mapping between information sets and nodes. 
Each node contains a hashmap of edges with key being a player's move. Each edge contains the statistics of the particular move and a pointer to the next node as observed in the current iteration of MCTS. The next node pointer might vary in subsequent iterations if the same move can have multiple outcomes (non-determinism) and thus lead to various information sets. From the PIMC concept, we borrow the idea of determinizations. At the beginning of each MCTS iteration, a copy of a hidden information state is determined into a perfect information state. This is not to be confused with information set. The default solution to determinization is to sample the state randomly among possible legal states. However, when generating games for machine learning experiments, we used the ``cheater'' approach that can determinize the correct state. Such an approach is often used in teaching sessions. In particular, in card games, human experts teach beginners how to play with open cards. In our case, the justification is that the ``cheater'' allows for generating stronger games quicker.\\

In our implementation, there are two interfaces for the concept of the game state:\\
\textbf{Game state for simulations (\textit{GS})} - this is the only interface used to apply the logic of the game such as determining legal moves, applying moves, checking whether the game has ended or getting the result of the game. This interface is used both in the selection and simulation phases. However, in the selection, the other interface (information sets) is used as well.\\
\textbf{Information Set Game state for statistics \textit{(IS})} - this is an abstraction of a state with possible hidden information. It represents all kind of information, based on which a player will take actions. The idea is to use only a subset of the simulation game state in order to group states. Such a separate interface not only allows for ignoring hidden information but also for reducing the resolution of the state. For instance, states that are similar in terms of some arbitrary measure can be grouped together. The information sets in our approach are plain data storage objects. The only methods the \textit{IS} interface contains are \textit{hash} and \textit{equals}, what enables efficient equality comparisons.

After the \textit{GS} has been determined, the selection phase starts from the root node. In each visited node during that phase, the set of currently legal moves is computed and intersected with the set of all moves observed in the node so far. 
Each move is associated with an edge. Active edges are the ones that correspond to moves that are currently available. The active edges are scored according to the selection formula (c.f. Equation~\ref{eq:uct}) and the best scored edge is chosen. Next, the \textit{GS} interface is used to apply the selected edge's move and compute the resulting state. This state is then used to generate an information set. We call this process capturing the information set and the \textit{GS} requires an implementation of the \textit{capture()} method that returns the \textit{IS} from a given player's perspective. The perspective is decided based on which player is active in the current state. Once the \textit{IS} is created, it is used to query the transposition table for the next node to traverse. If no such node exists, it is added to the transposition table with the key equal to the current \textit{IS} and the selection phase is terminated. 
The selection phase is repeated for the next node until the termination condition (a node visited for the first time) is not satisfied. Because nodes are matched with information sets, this statistics of actions performed within the same \textit{IS} are clustered together. Moreover, this allows to significantly reduce the combinatorial size of the game tree in comparison with using regular game states as nodes. When the selection phase ends, the last seen \textit{GS} is passed to the simulation phase as the starting state. The result of a simulation is propagated to all edges chosen in the selection phase.

\begin{algorithm}
\begin{algorithmic}[1]
\Procedure{iterate}{$state$}
\State $rootNode\gets $ \textbf{createRoot($state$)}
\State $node \gets rootNode$ \Comment{current node}
\While{$elapsedTime < allotedTime$}
\State $movingState\gets $ \textbf{determinize($state$)}
\While{$mcts.selection\not= finished$}
\If{$movingState.terminal \not= true$}
\State $node\gets node.select(movingState)$
\EndIf
\EndWhile
\State \textbf{propagate(simulation($movingState$))}
\EndWhile
\EndProcedure
\end{algorithmic}
\caption{Pseudocode of the main MCTS loop.\\
The $simulation$ method starts from the $movingState$ and performs a quasi-random simulation and returns the result of the game. It can be replaced by another evaluation procedure as discussed later in the paper.}
\end{algorithm}

\begin{algorithm}
\begin{algorithmic}[1]
\Procedure{node.select}{$movingState$}
\State $moves\gets movingState.getMoves()$
\State $currentEdges \gets []$
\For{\textbf{each} $move$ \textbf{in} $moves$}
\State $edge \gets allEdges[move]$ 
\If{$edge$ \textbf{not found}}
\State $edge \gets \textbf{new } edge(move)$
\State $allEdges[move] \gets edge$
\EndIf
\State $edge.N \gets +1$ \Comment{incr. observed count}
\State $currentEdges.push(edge)$
\EndFor
\State $chosenEdge\gets $ \textbf{selection($currentEdges$)}\Comment{UCT}
\State $chosenMove\gets chosenEdge.getMove()$
\State $chosenEdge.V\gets +1$ \Comment{incr. visit count}
\If{$chosenEdge.V == 1$}
\State $mcts.selection \gets finished$
\EndIf
\State $movingState.apply(chosenMove)$
\State $is \gets \textbf{capture}(movingState)$\Comment{create IS}
\State $tt \gets mcts.getTranspositionTable()$
\State $chosenEdge.nextNode \gets tt.findOrCreate(is)$
\State \textbf{return} $chosenEdge.nextNode$
\EndProcedure
\end{algorithmic}
\caption{Pseudocode of the inner MCTS loop. The $findOrCreate$ method accepts an information set and returns the corresponding node from the transposition table.}
\end{algorithm}

\subsubsection{Handling Randomness}

Non-determinism in games can quickly increase the combinatorial complexity to enormous levels. For example, there are $5.36 * 10^{28}$ different deals possible in the game of Bridge. 
Randomness is also prevalent in Hearthstone, with effects such as ``discover a random spell'' or ``deal from X to Y damage''. Each unique random outcome would most likely result in a different state, and therefore, would require its own node in the tree.

The novelty of our MCTS implementation is complete exclusion of nature moves. This makes the game modeling and simulating significantly easier using our library. Actions may include any non-determinism. This is possible, because we do not store game-states directly in the tree as results of actions. As shown on Algorithm (2), each time a move is played, we compute the resulting state dynamically, even if the move has been already sampled in previous iterations. The resulting state is used to create the information set, which then is used to fetch the next node to visit.
In consequence, statistics of moves are averaged according to the probability distribution of various random effects. If a move is good in average, the score will be high and it will be chosen more frequently in the selection phase of the MCTS algorithm.\\

\subsubsection{Handling Combinatorial Explosion}\label{boardsolver}

We have already introduced the idea of the separation of ``virtual game states'' modeled as Information Sets and the regular game states for simulations. This allowed us to gather statistics in a much more coarse-grained representation of state-space. 
However, the combinatorial complexity of the game is still very high due to the number of possible attacks, the fact that attacks can be done in chosen order and the options to intertwine playing cards between the attacks. The authors of~\cite{andersson2016programming} have calculated that, in the pessimistic case, 
there are approximately $10^{10}$ possible ways of performing the attacks. Quite often, however, lots of permutations of attacks will result in the same state in the end and there is no need to examine all of them. To tackle this problem, we have developed the so-called ``board solver'' - a heuristic that generates a sequence of attack actions in a given state. In general, the heuristic first checks if it can kill the opponent in one turn and does it if possible. If not, the heuristic will check whether the opponent is likely to win during their next turn and if so, the attacks will focus on killing the most threatening opponent minions. If no of these cases appear, the heuristic will score all possible single attacks based on the $gain - loss$ of the board potential. A single attack is a pair (attacker, defender). In Hearthstone, there are at most 8 attackers and 8 defenders, so, in the pessimistic case, 64 scores need to be calculated. The attacks are applied in a greedy fashion, i.e., the best scored attack is applied first (if possible), next the second best and the process continues until there are no more legal attacks. An application of an attack may render some of the following attacks illegal, for example when they use an attacker that has already attacked or defender that has already been killed. The heuristic for attacks is used as an artificial action in the game: ``use solver''. The MCTS is allowed to choose this action at any point during the turn, but only once per turn. Once the action is chosen, the attack moves are generated and applied, to there will not be any attacks move anymore during the turn for the minions that are already on the board.\\

\subsubsection{Interfacing heuristics with MCTS}

The MCTS algorithm is quite powerful on its own, but it can still benefit from domain-specific optimizations. It has been proven that, in more complex games such as Go~\cite{alphaGo} with huge branching factor and delayed rewards of taking actions, the vanilla method needs to be enhanced by some form of heuristics. 

This weakness has motivated us to combine this algorithm with heuristics represented by prediction models. Such prediction models can be trained to either predict the outcome of the game by looking at a potential next state (candidate state) of the game or at a potential action (candidate action). In the scope of this paper, we will use the terms ``machine learning prediction models'' and ``heuristic evaluation'' interchangeably.   

There is a couple of ways to combine external heuristics with the MCTS algorithm. The authors of paper~\cite{magician} give a nice review of four common methods: Tree Policy Bias, Simulation Policy Bias, Early Cutoff and Move Ordering. We use the first three of them: 

\textbf{(1) Tree Policy Bias} - here the heuristic evaluation function is included together with the $Q(s,a)$ in the UCT formula (see Eq.~\ref{eq:uct}) or its equivalent. A typical implementation of this idea is called \textit{Progressive Bias}~\cite{chaslot2008progressive}, in which the standard UCT evaluation is linearly combined with the heuristic evaluation with the weight proportional to the number of simulations. The more simulations are performed, the more statistical confidence, and therefore, the higher weight is assigned to the standard UCT formula.\\
 \indent \textbf{(2) Simulation Policy Bias} - here the heuristic values affect probabilities of certain actions in the simulation phase to make simulated players stronger and, therefore, each simulation a better approximation of a potential future game. The two most common implementations are pseudo-roulette selection with probabilities computed using Boltzmann distribution (where the heuristic evaluation is used) or the so-called epsilon-greedy approach~\cite{miniplayer-tciaig}. In the latter, the action with the highest heuristic evaluation is chosen with the probability of $\epsilon$ or a random one with the probability of $1-\epsilon$.\\
\indent \textbf{(3) Early Cutoff} - terminate the simulation earlier (e.g., with some probability or at fixed depth) and return the heuristic evaluation of the last reached state instead of the terminal one. In~\cite{magician}, this enhancement is reported to achieve the best results among the tested methods.

The aforementioned AlphaGo program employs both, Tree Policy Bias and Simulation Policy Bias. Motivated by its success, we decided to apply a similar approach for Hearthstone.

\section{Augmenting MCTS with Machine Learning}\label{sec:exps_performance}

The state of the art implementations of MCTS, such as AlphaZero, use deep neural networks for providing heuristic evaluations of states and actions. Two main approaches are used --  so called \textit{value network} is a deep neural network that provides the predictions of a game outcome given a state of the game. The predictions are usually provided as scores which can be interpreted as probabilities of winning the game by each player. Such predictions may be used by MCTS to foresee an outcome of a playout without simulating it until the terminal state, or even to entirely replace the simulation phase. A \textit{policy network} is another type of a neural network that given the state of a game provides values of each action available in that state. Policy network may thus provide information about which actions should be chosen in a state. As shown in \cite{alphaGo,chess_and_shogi}, the use of value and policy network heuristics significantly improves the performance of MCTS methods, enabling them to beat humans in very complex games.

In our solution we will focus on the value network heuristic for Hearthstone. We will use an iterative approach to neural network training, which uses large amount of hearthstone games, generated by self-playing bots.

\subsection{Game-state vectorization with embeddings}

Heuristic functions for evaluating game states require a vectorized representation of the state. It is common to use hand-crafted attributes to represent particular aspects of the state and then, using some weighted combination of those attributes, derive a value representing the utility of a state. While this approach works for games such as chess, it may be difficult to engineer such attributes for much more complex games such as Go or Hearthstone. As we use deep learning methods for obtaining heuristic functions, it is possible to represent Hearthstone states by large vectors composed of values of low-level features such as: attributes of each minion on the board (HP, attack, taunt, charge etc.), attributes of each player (HP, weapons, mana, hero type, etc.), attributes of cards in hand (type, mana cost, etc.) and general attributes (turn number, cards in deck, etc.). Moreover, as most cards in Hearthstone have custom descriptions that define special effects, it is necessary to extend the vectors by meaningful representations of particular cards.



One way to represent the cards in a relatively low-dimensional vector space is by using a word2vec model \cite{Mikolov:2013:DRW} to learn the embeddings from cards' textual descriptions. It can be done either by aggregating vector representations of words from the texts or by training a paragraph vector model \cite{Le:2014:DRS}, where each paragraph corresponds to a single card. Since descriptions of Hearthstone cards are relatively short and use a limited vocabulary, it is expected that a dimensionality of our embeddings should be much lower than in other common applications of the word2vec model. We experimentally checked that using more than 16 dimensions brings negligible improvements, and thus we used embedding size $10$ in our further experiments. To learn the embeddings, we used the skip-gram model implemented in TensorFlow. Apart from the embedding size, standard parameter values were used, i.e. context size was set to 10 and the batch size was 256. The model was trained for 300 epochs using a stochastic gradient descent optimizer, with a learning rate $0.1$, decreased by a factor of $10^{-1}$ after every 100 epochs.

In our final solution, we used a vectorizer that had 750 elements, including all low-level features for both players and utilized embeddings to represent all cards and minions.

\subsection{State evaluation with value network}

Our state evaluation heuristic uses a fully connected neural network for providing the win probabilities of each player. The network consists of three dense layers with 256, 128 and 64 neurons respectively and uses \textit{tanh} activation function. The input is a vector of size 750 (as described in the previous section), while the output consists of two neurons with a \textit{softmax} activation. The network thus solves a classification task: given a state predict the winner.

The training data for the network is generated by recording games played between bots. During a simulation, the state of the game is vectorized to vector $\vec{S}$ at each step, and the final score of the game is stored as a two-element vector: $ \vec{score} = [p1_{score}, p2_{score}]$. Next, the vectorized states are sampled randomly with some probability $p$ and pairs $[\vec{S}, \vec{score}]$ are added to the training dataset. Random sampling is required, as consecutive states are highly correlated. Finally the network is trained to provide score given a state vector. We used ADAM optimizer with learning rate = $0.001$

Value networks are trained to predict scores of games that were played with different decks as well as from the perspective of any of the two players. However, the accuracy of the predictions are better if there are separate networks trained for particular decks and even for particular player positions (first or second player). 

In our preliminary tests we created a dataset with over 3.5M samples from games played by strong MCTS bots (cheater MCTS with 1 second per move) playing with 400 different decks. The network were trained to predict outcomes of the games played with any of the available decks and for any of the players. The accuracy of the value network trained using this dataset was evaluated on a separate validation set and reached $0.76$.

We have used the trained value network for early termination of random simulations. The termination was done after the last move of a player in turn, but not earlier than after k=20 steps. After termination, the statistics in MCTS tree were updated with probabilities of winning obtained from the value network.

\subsection{Iterative learning - mastering Hearthstone}\label{continous_learning}

To further improve the performance of our solution, we have prepared an environment for continuous, iterated learning of our machine learning models. The main idea is that  MCTS with a heuristic may be used to generate games of progressively better quality. Those games may then be used to create more accurate heuristics, which may be used to generate games of even better quality. This process may be repeated many times for better optimization of the heuristics.

In our approach to iterative learning, we have started with plain MCTS to generate over 20000 games. Next, those games were used to generate an initial dataset consisting of randomly selected states and corresponding scores. Models for value networks were trained and used to generate the next version of the bot. Then, in each iteration, the bot played 3000 games, from which new state-score pairs were sampled and added to the training dataset. The training dataset length was clipped to 1M samples, so that after a few iterations older samples were removed and most recent samples were appended as in a FIFO buffer. The state-score pairs were sampled with probability $p=0.5$. In each iteration, value networks were retrained from scratch using 80\% of the training dataset. Remaining 20\% was used for validation of the network.

Using iterated learning, we were able to achieve an accuracy of 0.775 for the first player and 0.794 for the second player, when training for one type of deck only. In the next section we describe in details the performance of particular bots.





\section{Experiments}

\begin{table}[t]
\centering
\caption{Evaluation results - 0.5 second per move}	
\label{eval2}
\begin{tabular}{llllll}
\toprule
P1              & \thead{P1\\wins}  & \thead{P2\\wins} & \thead{P1\\win \%} & \thead{P2\\win \%} & P2 \\
\midrule
mcts    & 735     & 265     & 73,5\%    & 26,5\%    & mcts            \\
\midrule
mctsVS        & 500     & 0       & 100,0\%   &           & random          \\
mctsVS & 391     & 108     & 78,4\%    &           & mcts            \\
mctsV  & 410     & 90     & 82,0\%    &           & mcts            \\
mctsS  & 395     & 105     & 79,0\%    &           & mcts            \\
\midrule
random          & 0       & 500     &           & 100,0\%   & mctsVS        \\
mcts            & 219     & 280     &           & 56,1\%    & mctsVS \\
mcts            & 249     & 251     &           & 50,2\%    & mctsV \\
mcts            & 266     & 234     &           & 46,8\%    & mctsS \\
\bottomrule
\end{tabular}
\end{table}

\begin{table}[t]
\centering
\caption{Evaluation results - 1 second per move}
\label{eval1}
\begin{tabular}{llllll}
\toprule
P1              & \thead{P1\\wins}  & \thead{P2\\wins} & \thead{P1\\win \%} & \thead{P2\\win \%} & P2 \\
\midrule
mcts            & 705     & 294     & 70,6\%    & 29,4\%    & mcts            \\
\midrule
mctsVS        & 500     & 0       & 100,0\%   &           & random          \\
mctsVS & 364     & 135     & 72,9\%    &           & mcts            \\
mctsV  & 380     & 120     & 76,0\%    &           & mcts            \\
mctsS  & 358     & 143     & 71,6\%    &           & mcts            \\
\midrule
random          & 0       & 500     &           & 100,0\%    & mctsVS \\
mcts            & 224     & 276     &           & 55,2\%    & mctsVS \\
mcts            & 220     & 279     &           & 55,9\%    & mctsV \\
mcts            & 263     & 236     &           & 47,3\%    & mctsS \\
\bottomrule
\end{tabular}
\end{table}

\begin{table}[t]
\centering
\caption{A summary of results obtained in games between AI agents and human opponents.}	
\label{eval3}
\begin{tabular}{llllll}
\toprule
P1              & \thead{P1\\wins}  & \thead{P2\\wins} & \thead{P1\\win \%} & \thead{P2\\win \%} & P2 \\
\midrule
Regular            & 7     & 7     &           & 50\%   & mctsVS-1s \\
Legend            & 12     & 9     &           & 43\%    &  mctsVS-1s \\
mctsVS-1s         & 9     & 6     & 60\%      &     & Regular \\
mctsVS-1s         & 3     &15     & 17\%      &     & Legend \\
\bottomrule
\end{tabular}
\end{table}

We have conducted a series of experiments to measure the skill of various Hearthstone bots based on MCTS and different heuristics. Due to the high complexity of Hearthstone, mainly caused by the large number of possible decks and the impact of random effects on the game outcome, we have restricted our test cases to only two decks: \textit{ZooWarlock} and \textit{CubeWarlock}. Moreover, we have fixed the positions of both players, so that \textit{ZooWarlock} deck was always played by the first player, while \textit{CubeWarlock} by the second.

In order to obtain the best possible version of the value network, we have run iterative training for 64 iterations. Next, we have created a hearthstone bot for each version of the  value network obtained during the iterative learning. Finally, we have used 64 versions of the bot to play over 50k matches between themselves and assigned a glicko2 rating \cite{glicko} to each bot. Based on the glicko2 rating, we have selected the best bot, and thus the best value network, for the first and second player (obtained from 21st and 33rd iteration respectively).

For our final evaluation, we have compared plain MCTS (denoted by \textit{mcts}) with two different heuristics: a) previously selected, best value networks from iterative learning - denoted by~\textit{V}; b) board solver described in section \ref{boardsolver} - denoted by~\textit{S}. We have measured the impact of the value network, board solver and both of those combined together. Each configuration of the bot was used to play 500 games against plain MCTS bot. Moreover, we have also compared our solution with a randomly playing bot. To have a baseline for the performance, a 500-game match between only plain MCTS bots was played as well. The games were played with two time limits per move used by MCTS: 0.5 and 1.0 second. The results are presented in tables \ref{eval2} and \ref{eval1}. The strength of each bot is measured by the percentages of won games.

The baseline win-rates are 73.5\% for the first player and 26.5\% for the second in case of 0.5 second per move time limit. Increasing the time limit improves the strength of the second player, resulting in win-rates 70.6\% for the first player and 29.4\% for the second. The evaluation results show that each heuristic has a noticeable impact on the strength of the bot. As the first player has already a high win-rate, adding heuristics improves the win-rate by up to 9 percentage points. However, in case of the second player, adding heuristics may even double the win-rate. 

It is important to note here that the type of deck used has a huge impact on the strength of the bot. The deck used by the first player has an aggressive, but fairly straightforward, style of play. The deck used by the second player, has on the other hand, a lot of complex strategies and needs to be played carefully; yet used by a skillful player, it has a much greater winning potential compared to the first deck. This fact may help to understand why the strength of the second player is increased so dramatically when using well-crafted heuristics.

Moreover, heuristics provide a larger advantage, when playing with lower time per move limit as MCTS performs a fewer number of iterations. A combination of a value network and board solver, when only 0.5 seconds per move are available for the MCTS to perform simulations, provide the greatest boost to the bot's strength. With 1 second per move available, the difference between using only value network and the combination of value network and board solver is minimal.

Finally, we have arranged matches between a few hearthstone players and our bot. The results are presented in table \ref{eval3}. Games were played by two regular players (Hearthstone rank $>15$, which is held by approx. 75\% players) and two players with a Legendary rank (the best one with less than 0.5\% of players). 

\section{Conclusions}

In this paper, a fully-fledged approach to constructing a Hearthstone playing bot was presented. Some novel features of the approach include modification of the MCTS algorithm to handle randomness without explicitly defined nature moves, a combination of the PIMC and ISMCTS methods to tackle imperfect information, and a heuristic solver for calculating attacks in Monte Carlo simulations. In addition, we designed and conducted machine learning experiments aimed at learning game state evaluation functions. Finally, an iterative learning loop aimed at creating the ``ultimate bot'' was proposed.

We can conclude that the resulting agent is likely to be among the strongest Hearthstone bots at the moment. Although Hearthstone has become a testbed for AI, there has not been yet proposed any universal benchmarking methods, so it is difficult to assess the strength other than by human observation, self-play between various versions of the agent or a random player. However, in all cases, the proposed solution shows its upper hand. The bot is able to win, with an impressive consistency, 100\% games against the random player. It is also capable of winning games against Legend rank players, which alone can be regarded as very promising. The human players reported that in many situations they felt the bot played really well. Finally, we have shown the progressive improvement of the bot's skills by sparing it against previous versions. We designated two decks for this experiment, but the approach can be generalized for any number of decks easily, e.g., as an ensemble that chooses the right model (or even blends a few of them) for the deck on the fly.    

In order to benchmark our agent against other Hearthstone bots, we plan to submit it to the 2018 Hearthstone AI Competition held under the CIG (Computational Intelligence in Games) conference. Our submission to this competition will differ with the approach described in this paper in several details. It will work with the SabberStone (\url{https://github.com/HearthSim/SabberStone}) simulation engine as this is the official engine to be used during the competition. This simulator is only able to simulate approximately 200 games per second, on a modern high-end consumer PC, whereas our simulator performs 10000 games, on average. Because of this fact, we choose to limit the depth of the Monte Carlo simulations to the end of a single turn. At the end of the turn, the state evaluation function powered by machine learning will be used. We hope that the solutions adopted for the CIG competition will help us in designing even more cunning artificial Hearthstone agent, and as a consequence, move us one step further in the pursuit of the Grail of video games -- smarter and challenging AI.


\bibliography{refs}
\bibliographystyle{IEEEtran}

\end{document}